**COVER LETTER**

**Title of the manuscript**

General-Purpose vs. Domain-Adapted Large Language Models for Extraction of Structured Data from

Chest Radiology Reports

**Complete author list**

- Ali H. Dhanaliwala, University of Pennsylvania Department of Radiology

- Rikhiya Ghosh, Digital Technology and Innovation, Siemens Healthineers

- Sanjeev Kumar Karn, Digital Technology and Innovation, Siemens Healthineers

- Poikavila Ullaskrishnan, Digital Technology and Innovation, Siemens Healthineers

- Oladimeji Farri, Digital Technology and Innovation, Siemens Healthineers

- Dorin Comaniciu, Digital Technology and Innovation, Siemens Healthineers

- Charles E. Kahn, Jr. University of Pennsylvania Department of Radiology

**Description of any subject overlap with previously published works**

Not applicable

**Article Type**

Research and Applications

**Summary Statement**

A system using a pre-trained large language model (LLM) adapted for radiology (RadLing-System)

outperformed a dynamic few-shot prompting system using a general-purpose LLM (GPT-4-System) in extracting and standardizing relevant features from chest radiology reports (F1 score for extraction: 97% vs 78%, and standardization: 98% vs 94%, p<0.001).

**Key Results**

- In this retrospective study on extracting and standardizing features to Common Data Elements (CDE) from chest XR reports (900 training/499 test), a domain-adapted LM system (RadLing-System) was compared to a general-purpose LLM system (GPT-4-System).

- RadLing-System was more performant than GPT-4-System in feature extraction (F1 score 97% vs 78%), standardization via CDE mapping (98% vs 94%), and differentiating between *absent* (99% vs 64%) and *unspecified* (99% vs 89%).

- RadLing-System's light weight mapper further enables local deployment and lower runtime costs.

**TITLE PAGE**

**The title of the manuscript**

General-Purpose vs. Domain-Adapted Large Language Models for Extraction of Structured Data from Chest Radiology Reports

**The first and last names, middle initials, academic degrees, and institutions of all authors**

- Ali H. Dhanaliwala, MD, PhD, University of Pennsylvania Department of Radiology

- Rikhiya Ghosh, PhD, Digital Technology and Innovation, Siemens Healthineers

- Sanjeev Kumar Karn, PhD, Digital Technology and Innovation, Siemens Healthineers

- Poikavila Ullaskrishnan, Digital Technology and Innovation, Siemens Healthineers

- Oladimeji Farri, M.B.B.S, PhD, Digital Technology and Innovation, Siemens Healthineers

- Dorin Comaniciu, PhD, Digital Technology and Innovation, Siemens Healthineers

- Charles E. Kahn, Jr., MD, MS, Department of Radiology and Institute for Biomedical Informatics, University of Pennsylvania

**The name and street address of the institution from which the work originated**

1. University of Pennsylvania

Department of Radiology

3400 Spruce Street

Philadelphia, PA 19104

USA

2. Siemens Medical Solutions USA Inc,

755 College Road East,

Princeton, NJ 08540

USA

**The telephone number, e-mail address, and complete address (name, street address, postal or zip code) of the corresponding author**

Ali Haider Dhanaliwala, MD, PhD

Department of Radiology

University of Pennsylvania

3400 Spruce St

Philadelphia, PA 19104

Tel: 800-789-7366

[ali.dhanaliwala@pennmedicine.upenn.edu](mailto:ali.dhanaliwala@pennmedicine.upenn.edu)

**Word Count for Text**

3789

**STRUCTURED ABSTRACT**

**Objective**

Radiologists produce unstructured data valuable for clinical care when consumed by information systems. However, variability in style limits usage. This study compares performance of domain-adapted language model system (RadLing-System) to general-purpose large language model system (GPT-4-System) in extracting relevant features from chest radiology reports and standardizing to common data elements (CDEs).

**Materials and Methods**

Three radiologists annotated a retrospective dataset of 1399 chest XR reports (900 training, 499 test) to 44 unique pre-selected relevant CDEs (total of 21956 elements for the test set). RadLing-System involved parsing, retrieving and extracting values of CDEs from constituent sentences of a report sequentially while GPT-4-System was prompted with the report, features, value-set and dynamic few-shots to extract values and map to CDEs. The output key:value pairs were compared to reference standard at both stages. Identical match was considered true positive.

**Results**

F1 score for extraction was 97% (21296/21956) for RadLing-System and 78% (17125/21956) for GPT-4-System and for standardization was 98% (21296/21956) and 94% (20638/21956) respectively; difference was statistically significant (P<.001). RadLing-System demonstrated higher capabilities in differentiating between absent (99%(2993/3024) vs 64%(1935/3024)) and unspecified (99%(18120/18304) vs 89%(16290/18304)). RadLing-System's embeddings improved the extraction performance of GPT-4-System to 92%(16920/184) by giving more relevant few-shot prompts.

**Discussion**

RadLing-System's domain-adapted embeddings outperform dynamic few-shot general-purpose GPT-4-System in value extraction and its light-weight CDE mapper achieves higher F1 score than GPT-4-System in standardization. RadLing-System makes radiology data machine-readable and actionable and offers operational advantages including local deployment and reduced runtime costs.

**Conclusion**

The domain-adapted RadLing-System surpasses general-purpose GPT-4-System in extracting and standardizing common data elements from radiology reports, while being operationally advantageous.

**BACKGROUND AND SIGNIFICANCE**

Radiologists interpret images, generate annotations, and produce reports that are rich in diagnostic information and prognostic indicators. Such data are predominantly unstructured and not machine readable, which hinders their integration into healthcare information systems for monitoring, prediction, automated reporting, and decision-making[1]. This limitation also affects interoperability across systems. This study evaluated the performance of natural language processing models in extracting structured information from free-text reports of chest X-rays. Using radiology Common Data Elements (CDE)[2] as a framework, the study compared systems using two large language models (LLM): the GPT-4 general-purpose model and the RadLing domain-adapted model.

Maintained by the RSNA and ACR, radelement.org CDEs[3] standardize the description and permissible values of attributes of radiological findings such as size, location, and composition. By encapsulating properties like anatomical locations and measurements, CDEs make radiology data structured, normalized and machine-readable. CDEs come with unique identifiers and can be organized into sets for specific applications. For instance, a Pulmonary Nodule CDE set includes elements for nodule size, location, and composition (Table 1). CDEs thus offer a structured framework for radiology reporting to facilitate both human interpretation and machine analysis.

| CDE | Value |
|---|---|
| RDE1301_Composition | RDE1301.9_unspecified |
| RDE1302_Size | 3.0 |
| RDE1304_Location | RDE1304.1_left lung |
| RDE1717_Presence | RDE1717.1_present |

Table 1: RadElement.org CDE output for *A tiny 3 mm nonspecific nodule in the left lung base*

Efforts to extract actionable data from radiology reports have evolved from prescriptive linguistic algorithms[4,5] to machine learning and neural networks. Earlier methods struggled with language inconsistencies, while classical statistical approaches were limited by handcrafted features[6]. Modern pre-trained transformer architectures such as BERT[7] and its adaptations, RadBERT[8], PubMedBERT[9] and ClinicalRadioBERT[10], have set new benchmarks for text extraction, especially in radiology. Recently, large generative models like GPT-4 and Llama-2 that expedite task adaptation also have shown promise to generate a report's impression text from radiology findings[11], but remain underexplored in extracting structured data from radiology reports. Most attempts still result in reporting templates that contain non-machine-actionable free-text sections[12]. In the field of information retrieval, studies have also shown that GPT-4 performs well in extracting oncologic phenotypes from lung cancer CT reports[13] and better than a domain-specific NLP tool for extracting information from chest radiology reports[14].

**OBJECTIVE**

Our study aims to fill this research gap in using language models for extracting structured facts from radiology reports, which require the linking of multiple entities and modifiers and can be interpreted by humans and machines[15]. This study aims to compare feasibility and reliability of using domain-adapted versus general-purpose LLMs to enhance the utility of radiology data for both retrospective and real-time applications.

**MATERIALS AND METHODS**

This retrospective study utilized anonymized reports from two data sites under data-use agreements. A technical glossary of the terms in this paper is provided in appendix (Appendix-Section A.9)

**Problem Definition**

The intended system should take a report and map them to CDE:value pairs for all selected features. An effective classification system requires well-demarcated classes to reduce ambiguity. For example, a search for pleural effusion in RadElement.org yields three CDEs *(RDE1652_Presence(RDES254_Pleural Effusion), RDE398_Pleural fluid (RDES76_Pneumonia), RDE1605_Pleural effusion (RDES209_COVID-19 on Chest CT))*. Hence, 44 CDEs were pre-selected and corresponded to presence/absence/quantitative value of important findings in chest radiology (Appendix-Table A.3). The CDEs were chosen such that it either encapsulated the concept without preconditions or was the closest available one. Some concepts like *Scoliosis* didn't have CDEs as of August 2023 version and hence custom CDE codes *(RDESHS08_Scoliosis(RDESSHS_SHS Chest XR )* were created for evaluation. The default value for each CDE was either zero (quantitative) or unspecified (qualitative). We intend to minimize misinterpretations and want the model to indicate *I don't know* rather than suggesting absent/present without indisputable evidence. Hence, absent can only be assigned if there is enough evidence in the report to categorically assign a value *absent* to a feature. Else, *unspecified* is the assigned value. This was also the guideline adopted for annotation process.

**Dataset**

This study utilized 1,399 anonymized English-language reports of chest radiography (CXRs) annotated by three radiologists with 5, 10, and 13 years of experience. The dataset was split into 900 reports for training and 499 for testing. No validation set was used. Annotation tool was developed using Streamlit[16] (Appendix-Figure A.1).  Inter-reader variability was measured using Krippendorff's Alpha for multilabel problems. Reports were assigned randomly to annotators. There was a total of 21956 features to be extracted i.e., 44 features for each of the 499 reports with varying levels of feature frequency at sentence-level (Appendix-Figure A.2)

**RadLing-System**

RadLing-System comprised two phases: 1) Selection of feature class associated with the text, 2) CDE attribute and value mapping, which involves mapping the extracted features to relevant CDEs and their values from text. (Figure 1)

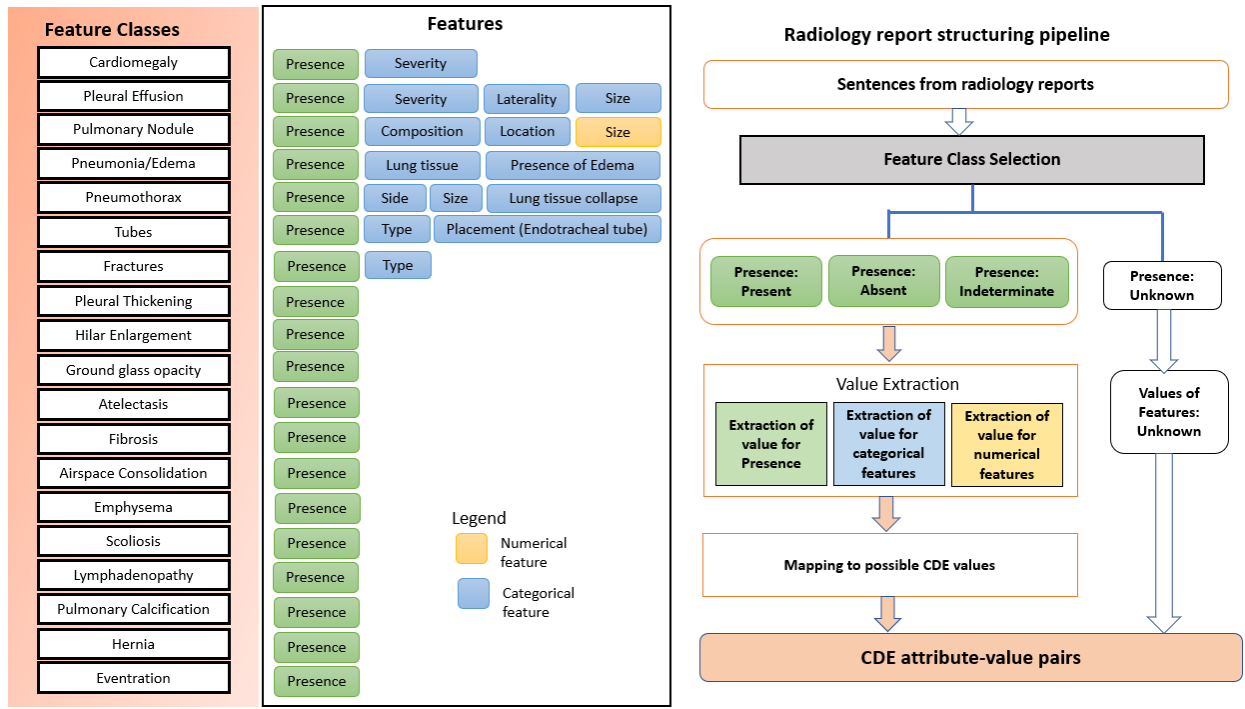

Figure 1: Caption: Domain-adapted Pre-trained Language Model System (RadLing-System) to convert radiology data to Common Data Elements (CDE) Alt-text: Diagram illustrating the RadLing System, a domain-adapted pre-trained language model designed to transform radiology data into standardized Common Data Elements (CDEs).

*RadLing: Radiology Domain-Adapted Pre-trained Language Model*

RadLing is a 335 million parameter domain-adapted language model (Figure 2) which outperforms RadBERT and PubMedBERT in standard radiology domain tasks[17]. Trained on a million radiology reports, it utilizes the transformer based discriminator-generator architecture of ELECTRA pretrained language model[18] and incorporates specialized vocabulary and radiology ontologies. During training, the model alternates between masking anatomical structures and clinical observations, enhancing its ability to interpret radiology-specific complexities.

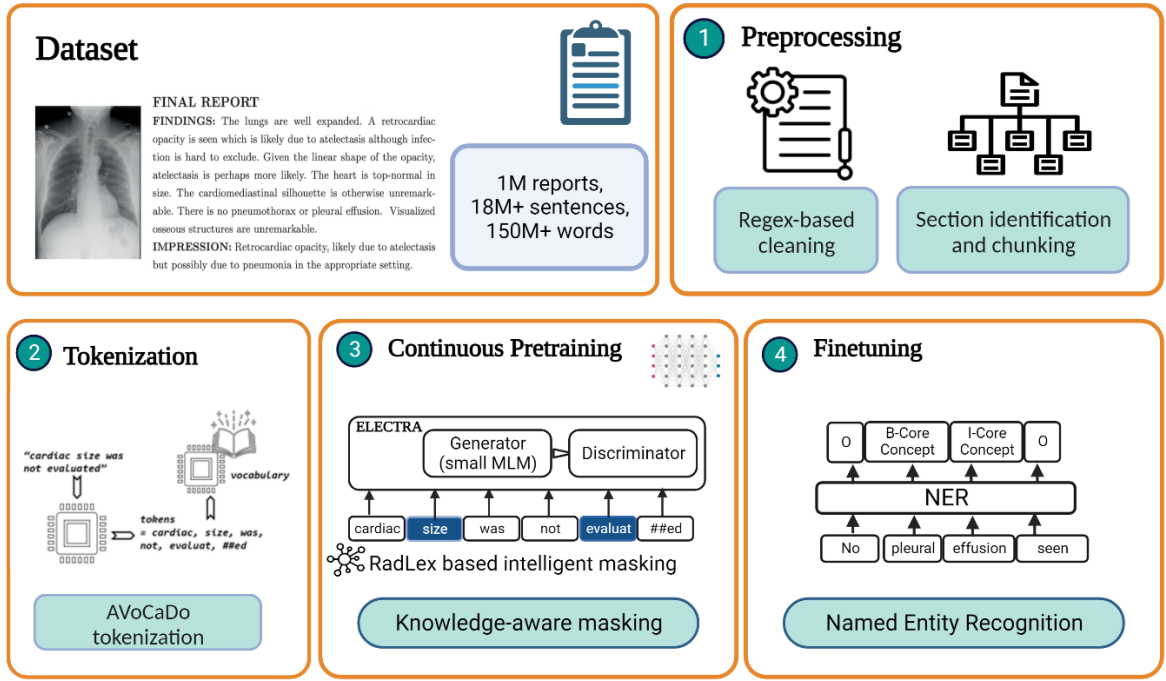

Figure 2: Caption: Schematic details of RadLing model architecture[17] including preprocessing, tokenization, continuous pretraining and fine tuning. RadLing model uses domain-specific vocabulary adaptation[19] using RadLex[20] and does a knowledge-aware masking for domain-adaptive continuous pretraining. We finetuned RadLing for Named Entity Recognition (NER) to find core concepts, locations and modifiers in a radiology report. Alt-text: Flowchart detailing the RadLing model's architecture, showing steps of preprocessing, tokenization, domain-specific vocabulary adaptation using RadLex, knowledge-aware masking for domain-adaptive pretraining, and fine-tuning for Named Entity Recognition (NER) to identify core concepts, locations, and modifiers in radiology reports.

*Sentence Transformer*

We trained a sentence transformer for RadLing model using Sentence-BERT architecture (SBERT)[21] with general domain semantic similarity (STS)[22] and Natural Language Inference (NLI)[23] datasets. Among the methodologies for generating sentence-level embeddings by utilizing encoder models[24-26], SBERT provides state of the art for encoder models in STS and NLI tasks. SBERT employs Siamese and triplet networks to train an encoder architecture to map input sentences into a fixed-size vector space with a contrastive learning objective that minimizes the distance between embeddings of similar sentence pairs (positive pairs) and maximizes the distance between embeddings of dissimilar sentence pairs (negative pairs). We modified RadLing model using SBERT to generate sentence embeddings.

*Selection of relevant feature classes from text*

The CDEs in the training data have been sorted into 19 related feature classes (Figure 1). The feature classes correspond to the commonly found abnormalities in a radiograph, and the various applicable attributes of these abnormalities comprise the CDEs. The attributes include and are not limited to presence, laterality, severity, composition, size and type.  Standard retrieval algorithm[27] combining lexical match followed by cross encoder vector comparison is used for this stage. The Findings and Impression section of the report are extracted and we apply retrieval and extraction algorithms sequentially on the sentences in the report. We perform lexical similarity using BM25[28], and we use stemmed entities extracted using RadLing from the constituent sentence of a report to be compared with the annotated sentences in the training set for each of the classes.  For a sentence in the test dataset, the candidate feature classes are selected based on lexical and semantic similarity matching with each of the core concepts with those in the sentence and those with higher than a threshold are selected as candidate feature classes. In cases where two sentences in a report are mapped to the same feature, we utilize values from the sentence which exhibits a higher semantic similarity with the feature in the report.

*CDE and value mapping*

Figure 1 shows the mapping of feature classes to features, most of which correspond to CDE attributes. The categorical values for the Type feature of feature classes Tubes (Types: Endotracheal tube, Central venous catheter, Pulmonary artery catheter, Nasogastric tube and Airway stent) and Fractures (Types: Acute Humerus Fracture, Acute Clavicle Fracture, Acute Rib fracture, Chronic rib fracture, Vertebral compression fracture) represent the corresponding CDEs. The value mapping pipeline involves extracting the associated values of these features in text and mapping them to the possible CDE and their values. It comprises of three steps: a) Extraction of value for the *Presence* attribute, b) Extraction of values for the categorical features and c) Extraction of values for the numerical features. The mapping of *Presence* attribute values is trivial since every CDE Presence attribute has the same values. Mapping of categorical values require semantic similarity using RadLing-System's semantic similarity module and numerical value mapping involves unit conversion and value range validation.

The *Presence* attribute can have four possible values: *Present*, *Absent* and *Unspecified*. For each feature that are not candidates for the sentence, the attribute value is assigned *Unspecified*. We have performed template-based data augmentation, as shown in Appendix-Table A.4, for the CDE values that do not have any annotation. RadLing-System's sentence similarity module is used to calculate cosine similarities between the embeddings of the sentence and the representative annotated training examples for the candidate feature classes to find the value with highest similarity. The categorical feature values are extracted using RadLing-System's NER to find locations and modifier elements from the sentences and then mapping them to the possible values in RadElements by semantic similarity to those of the annotated training examples and description of these values in the ontology. For numerical features, the algorithm searched for relevant numerical values using dependency tree parsing and checked against predetermined bounds as defined by RadElement.org. In addition, the units of these values are extracted and mapped to correct units by simple pre-defined rules. An example of RadLing-System in action for report is given in appendix (Appendix-Section A.7).

**GPT-4-System**

This system comprised the same two phases as RadLing-System but leveraged OpenAI models as of Aug 15, 2023. For the first phase of feature extraction, GPT-4 was given the original report, the 44 pre-selected features, its possible values and few examples of closest matching sentences as few-shot prompts. For few-shot prompting, the OpenAI text-embedding-ada-002 model[29] was used on each sentence in the original report and training dataset to generate embeddings. For each sentence in the original report, cosine similarity score was calculated against the training dataset and those with score above a threshold were added to prompt as few-shot examples (Appendix-Section A.8). This dynamic few-shot selection method has been demonstrated to efficiently steer generalist models like GPT-4 to achieve top performance, even when compared to models specifically finetuned for a specific domain[30]. For the second phase of standardization (mapping to CDEs) since the total number of CDEs in RadElement is 1060, embeddings-based approach was used instead of a prompt due to token limit of 8000 for GPT-4. Embeddings of the feature's name and value were compared with embeddings of each of the CDEs (including description, value set and CDE Set) with the highest cosine-similarity score being assigned the mapping.

**Statistical Analysis**

In our study, we conducted experiments to evaluate the effectiveness of our feature extraction and standardization process for radiology reports. The first phase (Extraction) involved the extraction of the 44 specific feature values from each report. The second phase (Standardization) focused on the mapping of extracted feature values to their corresponding Common Data Elements (CDEs) based on the RadElement CDE definitions.

F1 score measurement was done on a macro-averaging level to consider the imbalance of positive/absent/unspecified classes since *unspecified* had much larger instances. For each phase, they were calculated as:

- Extraction Phase: F1 score was assessed based on the match between the extracted feature value and the ground truth. For instance, an extracted value was deemed a true positive if it matched the ground truth value for that feature.

- Standardization Phase: The criterion for true positive ensured that both the feature and its associated value matched the ground truth CDE. This included correct identification and mapping to the specific RadElement CDE codes for the feature and value.

To illustrate, consider the statement "*The cardiac silhouette is enlarged*". If the ground truth for this observation was defined as "*RDE430_Cardiomegaly:RDE430.1_present*", the extraction phase aimed to correctly identify "*Cardiomegaly:Present*" as the feature-value pair. Subsequently, in the standardization phase, the correct outcome was "*Cardiomegaly:Present-RDE430:RDE430_1*", indicating a successful mapping to the appropriate CDEs. This two-phase approach allowed us to systematically evaluate the F1 score of feature extraction and standardization processes for both systems.

Fischer's exact test (python 3.9 *scipy==1.11.2)* was used to calculate p-value.

For GPT-4-System, the threshold for few-shot prompting was kept at 0.9 to only include highly similar few-shot examples. 0.9 was also the threshold used for feature class selection phase in RadLing-System.

**RESULTS**

For the 21,956 features in the 499 report test set, the F1 score, precision and recall for each stage is provided in Table 2. The Krippendorff's Alpha measuring inter-reader variability was 0.89.

| System | Extraction | | | Standardization | | |
|---|---|---|---|---|---|---|
| | Precision | Recall | F1 | Precision | Recall | F1 |
| GPT-4-System | 84% | 72% | 78% | 94% | 94% | 94% |
| RadLing- | 96% | 98% | 97% | 98% | 98% | 98% |

Table 2: Performance scores between RadLing-System and GPT-4-System in extracting and standardizing 21,956 features to 44 pre-selected Common Data Elements (CDEs) from 499 reports

Our results show that the RadLing-System performed better (F1 Score 97% vs 78%) than GPT-4 in the extraction and standardization phase (98% vs 94%) phases. Moreover, using the 2x2 results (Table 3) and *alpha* (for type 1 error) set at 0.05, we found *P<.001*.

| Criteria | Number of features |
|---|---|
| Both systems True | 19051 |
| Only GPT-4  system True | 48 |
| Only RadLing-System  True | 2867 |
| Both systems False | 390 |

Table 3: Cross-comparison of RadLing-System and GPT-4-System in extracting and standardizing 21956 features to 44 pre-selected Common Data Elements (CDEs)  to calculate p-value (*P<.001*)

The results for different classes (*positive, absent, unspecified*) in extraction phase is given in Table 4. The GPT-4-System demonstrated a varied performance across different classes in the context of feature extraction and standardization from radiology reports, with an overall F1 score of 78% across 21,956 instances. When zero-shot prompting was performed as opposed to dynamic few-shot prompting, the overall F1 score of GPT-4-System dropped to 36%. This is further evidenced by the low F1 score of 64% for *absent* (precision of 96% and recall of 47%) across 3,024 instances. The system's difficulty in distinguishing *'unspecified'* from *'absent'* suggests a potential area for improvement in handling ambiguous or insufficient data. The performance for *positive* class was also low with F1 score of 72%. This might indicate a need for enhancing the model's sensitivity or adjusting the training data to better capture positive cases. In comparison, the RadLing-System didn't demonstrate similar difficulty in differentiating between '*absent'* and '*unspecified'* values.

| Criteria | GPT-4 F1 score | RadLing F1 score | Instances |
|---|---|---|---|
| Overall F1 score | 78% | 98% | 21956 |
| F1 score for positive class | 72% | 95% | 628 |
| F1 score for absent class | 64% | 99% | 3024 |
| F1 score for unspecified class | 89% | 99% | 18304 |

Table 4: Comparative Accuracies of GPT-4-System and RadLing-System F1 score across various classes in extracting 44 pre-defined features from each report.

The per-feature scores are given in appendix (Appendix-Table A.6). RadLing-System performed worse than GPT-4 in 7 of the 44 features and achieved greater than 90% accuracy for 22 features. GPT-4-System performs better than RadLing-System for unseen CDEs or underrepresented CDEs in the training dataset. RadLing-System is unable to correctly determine the values of features like *Pulmonary calcification, Presence of Pneumoperitoneum, Subcutaneous emphysema* and *Volume of Pericardial effusion on CCTA* simply because they are not present in the training data. RadLing-System makes several mistakes for *unspecified* value for *Presence* of some of the features, like *Pleural Effusion, Chronic Rib fracture, Pneumonia*. For example, for  the sentence "*Mild blunting of the left costophrenic sulcus is suggestion of trace pleural fluid versus pleural thickening*.", *Presence* of *Pleural Effusion* is marked positive by RadLing-System and GPT-4-System against the ground truth *Unspecified*. More subtle possibilities of presence that are missed by RadLing-System include sentences like "*There is haziness of the lung*" which RadLing and GPT-4 systems determine as *Absence* of *Pneumonia* but should be

*unspecified* presence of *Pneumonia*. Other Presence-based errors by RadLing-System include "*There is no new consolidation*" which is marked as *unspecified* by RadLing-System but annotated by experts as absence of Airspace consolidation, or "*There is interval decrease in bilateral pleural effusions*" (RadLing-System: *unspecified*, Ground truth: *Present*). In both cases GPT-4-System performs better than RadLing-System. GPT-4-System performs worse for features like *Chronic rib fracture* (It marks "*There is chronic contour deformity of the lateral right fifth, sixth, and seventh ribs compatible with old, healed fractures*" as *unspecified* for Chronic rib fracture), *Cardiomegaly* (GPT-4-System unable to determine borderline or mild heart size increase as *Presence* of cardiomegaly), *Pneumonia* (understanding of pneumonia symptoms and *unspecified* presence of it) and identifying *Airspace consolidation* when mentioned in conjunction with *Atelectasis* (ex. "atelectasis/considation" or "atelectasis vs consolidation"). For other categorical features like *Laterality of Pleural Effusion*, RadLing-System has issues with 'bilateral' values, especially when this is implied by separately mentioning both left and right pleural effusion.

For the standardization phase, the embedding approach yielded F1 score rate of 94% for GPT-4-System. This success rate underscores the effectiveness of using embeddings to navigate the CDE space, ensuring that features are accurately standardized against the correct codes. The F1 score for value-to-CDE mapping reached 100%. This perfect score was facilitated by the relatively smaller set of possibilities—one value against four potential codes for each CDE, each with a descriptive element that guided the matching process. A notable issue encountered was related to the inherent duplication within the RadElement definitions. For instance, the feature "*Presence_Pleural_Effusion*" was erroneously mapped by GPT-4-System to "*RDE1605 - Pleural effusion*" in *RDES209 - COVID-19 on Chest CT*, rather than the more accurate "*RDE1652 - Presence*" CDE within the *RDES254 - Pleural Effusion* set. Such errors highlight the complexities of dealing with a rich and nuanced database like RadElement, where subtle distinctions between similar terms can lead to mismatches.

**Infrastructure Requirements**

The RadLing pre-trained model took 92 hours to train on Tesla V100 SXM2 machines with 8 GPUs and 16 GB memory per GPU (Appendix-Table A.5). The sentence-transformer stage took 8 hours to train on 4 such GPUs. There was no further training needed for the feature extraction and value mapping stages. For runtime, the entire RadLing-System took 10.89±2.16 seconds per report (~ 6 sentences or 98 tokens per findings section) when deployed on a Windows 10 Enterprise OS with Intel(R) Core(TM) i7-8850H CPU @ 2.60GHz processor and 16 GB RAM. GPT-4-System has ~1.7 trillion parameters and took around 13.42±1.80 seconds per report while deployed on a dedicated Azure OpenAI instance. Infrastructure requirements for running both the GPT-4 and text-embedding-ada-002 models on such an instance on cloud have not yet been provided by Microsoft. Another public LLM, Llama-2 with 70 billion parameters, requires at least two RTX 3090 GPUs with 16GB VRAM to run with comparable speed.

**DISCUSSION**

In this retrospective study on extracting normalized and machine-readable common data elements (CDE) from chest XR radiography reports, a domain-adapted language model system outperformed a general-purpose one with a F1 score of 97% (vs 78%) for extraction and 98% (vs 94%) for standardization to CDEs (P<.001). The domain-adapted system also demonstrated much higher F1 score in differentiating between *absent* (99% vs 64%) and *unspecified* (99% vs 89%).

The RadLing-System offers deployment flexibility and cost-effectiveness compared to large LLMs like GPT-4-System. It can run on local CPUs, enhancing data privacy, eliminating the need for high-performance GPUs and cloud deployments. For the specific task of feature extraction, the RadLing-System utilized a data augmentation approach using ground truth and RadElement.org definitions for features with less data, rather than comprehensive retraining. For deploying in other English-speaking sites that focus on similar imaging modalities and body regions, the existing model configuration possibly requires good quality data and finetuning to achieve comparable extraction performance.

When the embedding model used for few-shot prompting in GPT-4-System was replaced with RadLing-System's domain-adapted embeddings, the modified system achieved an F1 score of 92% for the feature extraction phase. Domain-adapted embeddings could be better performing than general-purpose embeddings in identifying correct data contexts for radiology tasks. Furthermore, the CDE mapping method employed in the RadLing-System that utilized dependency trees and semantic similarity techniques had better F1 score in CDE assignment than GPT-4 (98% vs 94%). Not only is this approach computationally more lightweight, faster, and cost-effective than using the very large GPT-4 for matching to CDEs, it also achieves higher scores. This makes the RadLing-System's domain-adapted capabilities a compelling choice for radiology applications, where both accuracy and efficiency are critical.

This study proposes a system that converts radiological text into structured format based on CDEs. This standardization makes data machine-readable and actionable, facilitating automated tasks like monitoring, reporting and decision-making in healthcare. This system could serve as a middle data-flow layer in clinical workflows, interacting with analytics engines and various translator AI models. Moreover, well-structured prompts have proven to enhance LLM performance in a variety of downstream tasks [31,32]. When general-purpose LLMs are employed in clinical settings for generative tasks, using structured CDE-formatted data as input can improve the F1 score and reliability of the model's output. This addresses issues like data hallucinations and increases the system's overall credibility.

Our study on the efficacy of large language models in structuring radiology data has limitations. It focuses on chest radiography in English, limiting generalizability. The evaluation used only 44 pre-defined CDEs and a small cohort of radiologists for annotations. The controlled setting doesn't capture real-world clinical complexities such as report addenda, errata, or the presence of ambiguous or conflicting statements. OpenAI models and RadElement.org CDEs were used as of August 2023, without subsequent updates. We also only benchmarked against a GPT-4-System using OpenAI embedding model. Systems using other LLMs like Llama-2, other approaches like fine-tuning or other public embedding models require further evaluation.

**CONCLUSION**

The study compared a radiology-adapted pre-trained language model system (RadLing system) with a system using general-purpose embeddings and LLMs (GPT-4-System) for extracting Common Data Elements (CDEs) from radiology reports. Results showed (P<.001) that the RadLing-System was more performant than GPT-4-System in feature extraction (97% vs 78%) and its light-weight mapper had better F1 score in standardization via CDE mapping (98% vs 94%). The RadLing-System also demonstrated higher capabilities in differentiating between *absent* (99% vs 64%) and *unspecified* (99% vs 89%) categories. Accuracy greater than 90% was achieved by RadLing-System for 22 of the 44 pre-selected features. RadLing-System's domain-adapted embeddings helped improve the performance of the GPT-4-System by giving more relevant few-shot prompts. RadLing-System's operational advantage includes local deployment and relatively less re-training requirements. Overall, the study presents a promising advancement in transforming radiology reports into actionable, structured data through the use of domain-adapted language models, with RadLing-System offering distinct advantages in both performance and operational efficiency.

**DISCLAIMER**

**ACKNOWLEDGMENTS**

We extend our gratitude to radiologists Dr. Pooja Hrushikesh Garud, Dr. Mihai Pomohaci, and Dr. Vishwanath R S, for meticulous annotation of over 1300 reports and valuable guidance in the design of the annotation tool.

Special thanks go to our supporting annotators at Siemens Healthineers: Rakshitha M, Prathap R, Barath Panneerselvam, Ivin Jose, Vini Jacob, Ashok Kumar, Deepa Anand, Saranya Krishnan, and Bindushree V, for their dedication and hard work.

We are indebted to employees in the Big Data Office at Siemens Healthineers for facilitating the procurement of data from collaboration sites. Our appreciation also extends to the Supercomputing team at Siemens Healthineers for overseeing the infrastructure essential for model training.

Finally, we are grateful to various other employees In Siemens Healthineers for their role in guiding the research study, aiding in data preprocessing, and contributing to the model-building system. Your collective contributions have been invaluable in the realization of this project.

**CONFLICT OF INTEREST**

This study was supported by Siemens Medical Solutions USA Inc. The authors have no financial relationship with Siemens Medical Solutions USA Inc that could be construed as a potential conflict of interest with respect to the research, authorship, and/or publication of this article.

**DATA SHARING STATEMENT**

 The data used in this study is proprietary and cannot be shared externally due to binding data usage agreements.

**REFERENCES**

1 Jorg T, Heckmann JC, Mildenberger P, Hahn F, Düber C, Mildenberger P, Kloeckner R, Jungmann F. Structured reporting of CT scans of patients with trauma leads to faster, more detailed diagnoses: An experimental study. *Eur J Radiol* 2021; 144:109954.

2 Rubin DL, Kahn CE Jr. Common data elements in radiology. *Radiology* 2017; 283:837-44.

3 Common Data Elements (CDEs) for Radiology. Available online: https://radelement.org/. Accessed October 21, 2023.

4 Wang Y, Mehrabi S, Sohn S, Atkinson EJ, Amin S, Liu H. Natural language processing of radiology reports for identification of skeletal site-specific fractures. *BMC Med Informatics and decision making*. 2019 Apr;19:23-9.

5 Rink B, Roberts K, Harabagiu S, Scheuermann RH, Toomay S, Browning T, Bosler T, Peshock R. Extracting actionable findings of appendicitis from radiology reports using natural language processing. *Proc AMIA Summits on Translational Science Proceedings*. 2013;2013:221.

6 Hassanpour S, Langlotz CP. Information extraction from multi-institutional radiology reports. *Artif Intell Med* 2016;66:29-39.

7 Kenton JD, Toutanova LK. BERT: Pre-training of Deep Bidirectional Transformers for Language Understanding. In *Proceedings of NAACL-HLT* 2019 (pp. 4171-4186).

8 Yan A, McAuley J, Lu X, Du J, Chang EY, Gentili A, Hsu CN. RadBERT: Adapting transformer-based language models to radiology. *Radiol Artif Intell* 2022; 4:e210258.

9 Gu Y, Tinn R, Cheng H, Lucas M, Usuyama N, Liu X, Naumann T, Gao J, Poon H. Domain-specific language model pretraining for biomedical natural language processing. *ACM Transactions on Computing for Healthcare (HEALTH)*. 2021 Oct 15;3(1):1-23.

10 Rezayi S, Dai H, Liu Z, Wu Z, Hebbar A, Burns AH, Zhao L, Zhu D, Li Q, Liu W, Li S. Clinical radiobert: Knowledge-infused few shot learning for clinical notes named entity recognition. *International Workshop on Machine Learning in Medical Imaging* 2022 Sep 18 (pp. 269-278). Cham: Springer Nature Switzerland.

11 Liu Z, Li Y, Shu P, Zhong A, Yang L, Ju C, Wu Z, Ma C, Luo J, Chen C, Kim S. Radiology-Llama2: Best-in-Class Large Language Model for Radiology. *arXiv* preprint arXiv:2309.06419. 2023 Aug 29.

12 Adams LC, Truhn D, Busch F, Kader A, Niehues SM, Makowski MR, Bressem KK. Leveraging GPT-4 for post hoc transformation of free-text radiology reports into structured reporting: a multilingual feasibility study. *Radiology*. 2023 Apr 4;307(4):e230725.

13 Fink MA, Bischoff A, Fink CA, Moll M, Kroschke J, Dulz L, Heußel CP, Kauczor HU, Weber TF. Potential of ChatGPT and GPT-4 for data mining of free-text CT reports on lung cancer. *Radiology*. 2023 Sep 19;308(3):e231362.

14 Savage CH, Park H, Kwak K, Smith AD, Rothenberg SA, Parekh VS, Doo FX, Yi PH. General-Purpose Large Language Models Versus a Domain-Specific Natural Language Processing Tool for Label Extraction From Chest Radiograph Reports. *American Journal of Roentgenology*. 2024 Jan 17.

15 Steinkamp JM, Chambers C, Lalevic D, Zafar HM, Cook TS. Toward complete structured information extraction from radiology reports using machine learning. *Journal of digital imaging*. 2019 Aug 15;32:554-64.

16 Streamlit A faster way to build and share data apps. https://streamlit.io/. Accessed October 21, 2023.

17 Ghosh R, Farri O, Karn SK, Danu M, Vunikili R, Micu L. RadLing: Towards Efficient Radiology Report Understanding. In: *Proceedings of the 61st Annual Meeting of the Association for Computational*

*Linguistics* (Volume 5: Industry Track); 2023. p. 640-651. Toronto, Canada: Association for Computational Linguistics.

18 Clark K, Luong MT, Le QV, Manning CD. ELECTRA: Pre-training Text Encoders as Discriminators Rather Than Generators. In *International Conference on Learning Representations* 2019 Sep 25.

19 Hong J, Kim T, Lim H, Choo J. AVocaDo: Strategy for Adapting Vocabulary to Downstream Domain. In *Proceedings of the 2021 Conference on Empirical Methods in Natural Language Processing* 2021 Nov (pp. 4692-4700).

20 Langlotz CP. RadLex: a new method for indexing online educational materials. *Radiographics* 2006; 26:1595-1597.

21 Reimers N, Gurevych I. Sentence-BERT: Sentence Embeddings using Siamese BERT-Networks. In *Proceedings of the 2019 Conference on Empirical Methods in Natural Language Processing and the 9th International Joint Conference on Natural Language Processing (EMNLP-IJCNLP)* 2019 Nov (pp. 3982-3992).

22 Cer D, Diab M, Agirre EE, Lopez-Gazpio I, Specia L. SemEval-2017 Task 1: Semantic Textual Similarity Multilingual and Cross-lingual Focused Evaluation. In *The 11th International Workshop on Semantic Evaluation* (SemEval-2017) 2017 Aug 3 (pp. 1-14).

23 Williams A, Nangia N, Bowman SR. A Broad-Coverage Challenge Corpus for Sentence Understanding through Inference. In *Proceedings of NAACL-HLT* 2018 (pp. 1112-1122).

24 May C, Wang A, Bordia S, Bowman SR, Rudinger R. On Measuring Social Biases in Sentence Encoders. In *Proceedings of NAACL-HLT* 2019 (pp. 622-628).

25 Zhang T, Kishore V, Wu F, Weinberger KQ, Artzi Y. BERT Score: Evaluating Text Generation with BERT. In *International Conference on Learning Representations* 2019 Sep 25.

26 Humeau S, Shuster K, Lachaux MA, Weston J. Poly-encoders: Architectures and Pre-training Strategies for Fast and Accurate Multi-sentence Scoring. In *International Conference on Learning Representations* 2019 Sep 23.

27 Thakur N, Reimers N, Rücklé A, Srivastava A, Gurevych I. BEIR: A Heterogeneous Benchmark for Zero-shot Evaluation of Information Retrieval Models. In *Thirty-fifth Conference on Neural Information Processing Systems Datasets and Benchmarks Track* (Round 2) 2021 Aug 29.

28 Robertson, Stephen, Hugo Zaragoza, and Michael Taylor. "Simple BM25 extension to multiple weighted fields." *Proceedings of the thirteenth ACM international conference on Information and knowledge management*. 2004.

29 OpenAI, 2023. Openai models overview. https://platform.openai.com/docs/models/overview. Accessed: August 25, 2023.

30 Nori H, Lee YT, Zhang S, Carignan D, Edgar R, Fusi N, King N, Larson J, Li Y, Liu W, Luo R. Can Generalist Foundation Models Outcompete Special-Purpose Tuning? Case Study in Medicine. *Medicine*. 2023 Nov;84(88.3):77-3.

31 Zhao Z, Wallace E, Feng S, Klein D, Singh S. Calibrate before use: Improving few-shot performance of language models. *International Conference on Machine Learning* 2021 Jul 1 (pp. 12697-12706). PMLR.

32 Pitis S, Zhang MR, Wang A, Ba J. Boosted Prompt Ensembles for Large Language Models. *arXiv* preprint arXiv:2304.05970. 2023 Apr 12.

**APPENDIX**

**Figures**

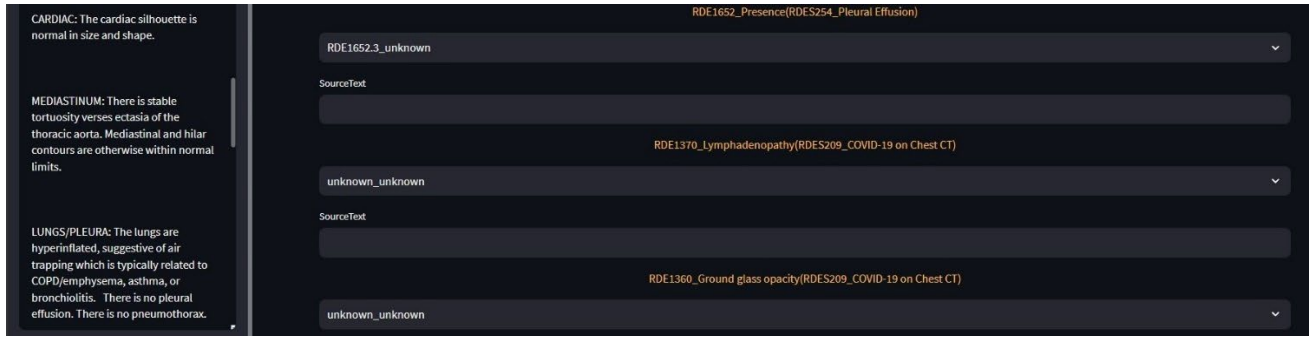

Figure A.1: Screenshot of annotation tool built using Streamlit platform ( https://streamlit.io/ )

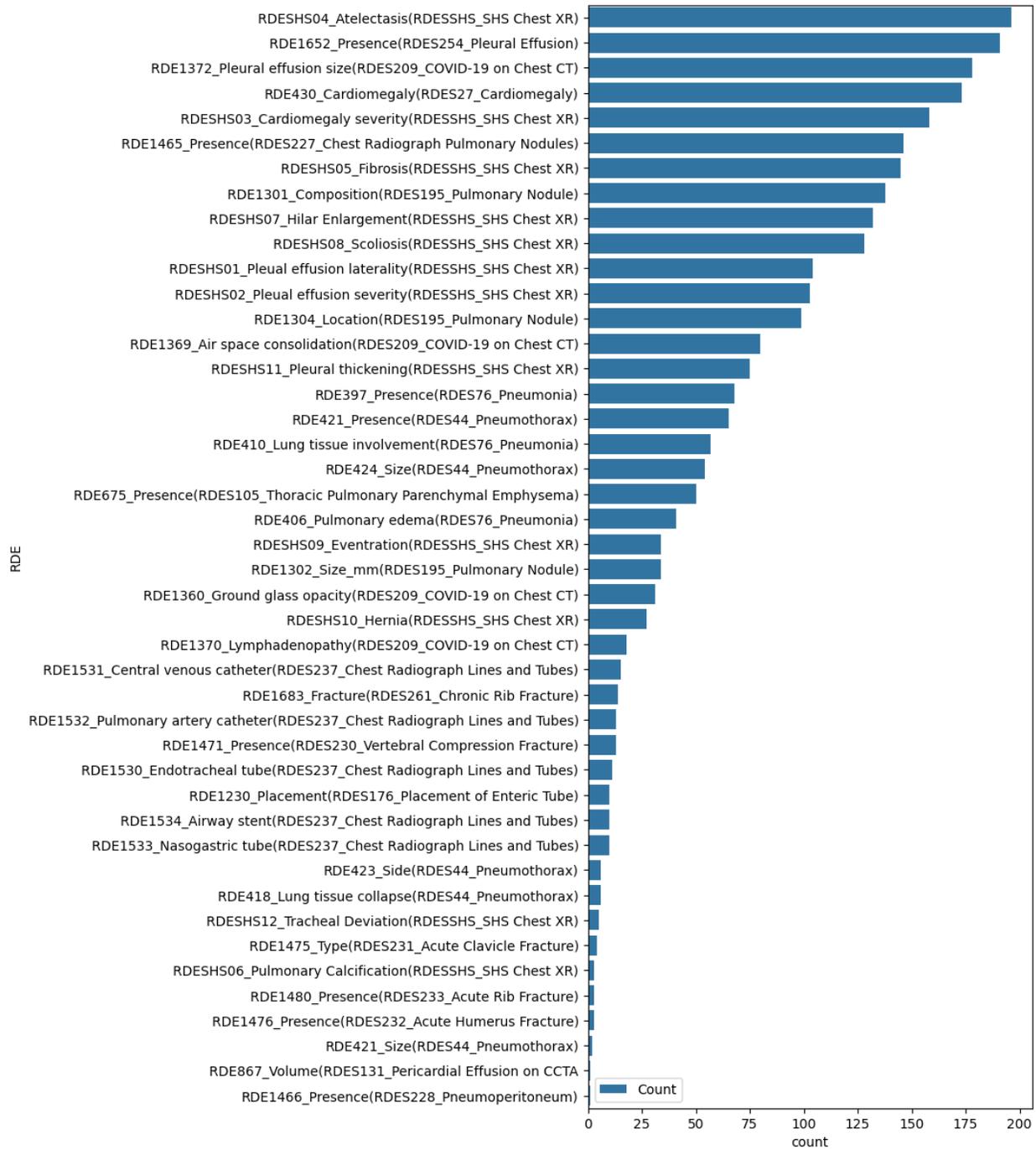

Figure A.2. Caption: Unique sentence count by Common Data Element (CDE) in training dataset of 900 reports. Alt-text: Bar chart showing counts of unique sentences for different Common Data Elements (CDEs) in a dataset of 900 radiology reports

**Tables**

| CDE | CDE Set |
|-----|---------|
| RDE430_Cardiomegaly | RDES27_Cardiomegaly |
| RDE1652_Presence | RDES254_Pleural Effusion |
| RDE1370_Lymphadenopathy | RDES209_COVID-19 on Chest CT |
| RDE1360_Ground glass opacity | RDES209_COVID-19 on Chest CT |
| RDE1369_Air space consolidation | RDES209_COVID-19 on Chest CT |
| RDE421_Presence | RDES44_Pneumothorax |
| RDE867_Volume | RDES131_Pericardial Effusion on CCTA |
| RDE406_Pulmonary edema | RDES76_Pneumonia |
| RDE1530_Endotracheal tube | RDES237_Chest Radiograph Lines and Tubes |
| RDE1531_Central venous catheter | RDES237_Chest Radiograph Lines and Tubes |
| RDE1532_Pulmonary artery | RDES237_Chest Radiograph Lines and Tubes |
| RDE1533_Nasogastric tube | RDES237_Chest Radiograph Lines and Tubes |
| RDE1534_Airway stent | RDES237_Chest Radiograph Lines and Tubes |
| RDE1230_Placement | RDES176_Placement of Enteric Tube |
| RDE1465_Presence | RDES227_Chest Radiograph Pulmonary |
| RDE397_Presence | RDES76_Pneumonia |
| RDE1475_Type | RDES231_Acute Clavicle Fracture |
| RDE1476_Presence | RDES232_Acute Humerus Fracture |
| RDE1480_Presence | RDES233_Acute Rib Fracture |
| RDE1683_Fracture | RDES261_Chronic Rib Fracture |
| RDE1471_Presence | RDES230_Vertebral Compression Fracture |
| RDE1372_Pleural effusion size | RDES209_COVID-19 on Chest CT |
| RDESHS01_Pleural effusion laterality | RDESSHS_SHS Chest XR |
| RDESHS02_Pleural effusion severity | RDESSHS_SHS Chest XR |
| RDESHS03_Cardiomegaly severity | RDESSHS_SHS Chest XR |
| RDE424_Size | RDES44_Pneumothorax |
| RDE423_Side | RDES44_Pneumothorax |
| RDE418_Lung tissue collapse | RDES44_Pneumothorax |
| RDE410_Lung tissue involvement | RDES76_Pneumonia |
| RDE1301_Composition | RDES195_Pulmonary Nodule |
| RDE1302_Size_mm | RDES195_Pulmonary Nodule |
| RDE1304_Location | RDES195_Pulmonary Nodule |
| RDESHS04_Atelectasis | RDESSHS_SHS Chest XR |
| RDESHS05_Fibrosis | RDESSHS_SHS Chest XR |
| RDESHS06_Pulmonary Calcification | RDESSHS_SHS Chest XR |
| RDESHS07_Hilar Enlargement | RDESSHS_SHS Chest XR |
| RDESHS08_Scoliosis | RDESSHS_SHS Chest XR |
| RDESHS09_Eventration | RDESSHS_SHS Chest XR |

| RDESHS10_Hernia | RDESSHS_SHS Chest XR |
| RDESHS11_Pleural thickening | RDESSHS_SHS Chest XR |
| RDESHS12_Tracheal Deviation | RDESSHS_SHS Chest XR |
| RDESHS13_Subcutaneous emphysema | RDESSHS_SHS Chest XR |
| RDE1466_Presence | RDES228_Pneumoperitoneum |
| RDE675_Presence | RDES105_Thoracic Pulmonary Parenchymal Emphysema |

Table A.3: 44 pre-selected relevant Common Data Elements (CDE) for chest radiology reports for

annotation and evaluation

| CDE | Example Augmented Sentences |
| --- | --- |
| RDE1530_Endotracheal tube | Endotracheal tube is adequately positioned, Endotracheal tube is absent, Endotracheal tube is malpositioned, Endotracheal tube is suboptimally positioned. |
| RDE1531_Central venous catheter | Central venous catheter is adequately positioned, Central venous catheter is absent, Central venous catheter is malpositioned, Central venous catheter is suboptimally positioned. |
| RDE1532_Pulmonary artery catheter | Pulmonary artery catheter is adequately positioned, Pulmonary artery catheter is absent, Pulmonary artery catheter is malpositioned, Pulmonary artery catheter is suboptimally positioned. |
| RDE1533_Nasogastric tube | Nasogastric tube is adequately positioned, Nasogastric tube is absent, Nasogastric tube is malpositioned, Nasogastric tube is suboptimally positioned. |
| RDE1534_Airway stent | Airway stent is adequately positioned, Airway stent is absent, Airway stent is malpositioned, Airway stent is suboptimally positioned. |
| RDE1230_Placement (RDES176_Placement of Enteric Tube) | Placement of enteric tube is correct, Placement of enteric tube is correct, Placement of enteric tube is misplaced, Placement of enteric tube is indeterminant, Placement of enteric tube is unspecified. |
| RDE1475_Type (RDES231_Acute Clavicle Fracture) | Acute clavicle fracture is proximal to sternum, Acute clavicle fracture is proximal to AC joint, Acute clavicle fracture is middle of the bone, between the sternum and AC joint, Acute clavicle fracture is unspecified, Acute clavicle fracture is absent, Acute clavicle fracture is unspecified. |
| RDE1476_Presence (RDES232_Acute Humerus Fracture) | Acute humerus fracture is present, Acute humerus fracture is absent, Acute humerus fracture is unspecified, Acute humerus fracture is unspecified |
| RDE1683_Fracture (RDES261_Chronic Rib Fracture) | Chronic rib fracture is present, Chronic rib fracture is absent, Chronic rib fracture is unspecified, Chronic rib fracture is unspecified |
| RDE1471_Presence (RDES230_Vertebral Compression Fracture) | Vertebral compression fracture is present, Vertebral compression fracture is absent, Vertebral compression fracture is unspecified, Vertebral compression fracture is unspecified |

Table A.4: Example augmented sentences for under-represented CDEs in RadLing-System's CDE identification step.

| Model Details | Values |
|---|---|
| RadLing Hyperparameters | learning rate 3e-5, AdamW optimizer and polynomial decay schedule with warmup, 235K steps |
| RadLing Sentence Transformer Hyperparameters | Warmupcosine scheduler, 10 epochs |
| RadLing Sentence Transformer embedding size | 1x768 |

Table A.5: Hyperparameter and vector size details of RadLing-System

| Feature | GPT-4 F1 Score | RadLing F1 score |
|---|---|---|
| Placement_Placement_of_Enteric_Tube | 1 | 1 |
| Presence_Acute_Humerus_Fracture | 0.16 | 1 |
| Side_Pneumothorax | 0.57 | 1 |
| Size_mm_Pulmonary_Nodule | 0.95 | 1 |
| Type_Acute_Clavicle_Fracture | 0.29 | 1 |
| Cardiomegaly | 0.95 | 0.99 |
| Presence_Thoracic_Pulmonary_Parenchymal_Emphysema | 0.99 | 0.99 |
| Size_Pneumothorax | 0.41 | 0.99 |
| Air_space_consolidation | 0.29 | 0.98 |
| Cardiomegaly_severity | 0.58 | 0.98 |
| Location_Pulmonary_Nodule | 0.81 | 0.98 |
| Lymphadenopathy | 0.78 | 0.98 |
| Presence_Pleural_Effusion | 0.93 | 0.98 |
| Presence_Pneumothorax | 0.95 | 0.98 |
| Hernia | 0.48 | 0.97 |
| Scoliosis | 0.49 | 0.97 |
| Airway_stent_Chest_Radiograph_Lines_and_Tubes | 0.88 | 0.93 |
| Presence_Chest_Radiograph_Pulmonary_Nodules | 0.59 | 0.93 |
| Nasogastric_tube | 0.88 | 0.91 |
| Pleural_effusion_laterality | 0.64 | 0.91 |
| Atelectasis | 0.56 | 0.9 |
| Central_venous_catheter | 0.89 | 0.9 |
| Composition_Pulmonary_Nodule | 0.78 | 0.9 |
| Hilar_Enlargement | 0.91 | 0.9 |
| Pleural_thickening | 0.56 | 0.9 |
| Endotracheal_tube | 0.88 | 0.89 |
| Pleural_effusion_size | 0.96 | 0.89 |
| Fracture_Chronic_Rib_Fracture | 0.18 | 0.88 |
| Pleural_effusion_severity | 0.63 | 0.87 |
| Pulmonary_edema | 0.65 | 0.86 |
| Ground_glass_opacity | 0.78 | 0.83 |
| Lung_tissue_involvement_Pneumonia | 0.65 | 0.81 |
| Eventration | 0.45 | 0.8 |
| Fibrosis | 0.5 | 0.8 |
| Lung_tissue_collapse_Pneumothorax | 0.58 | 0.8 |
| Tracheal_Deviation | 0.5 | 0.8 |
| Pulmonary_artery_catheter | 0.88 | 0.78 |
| Presence_Vertebral_Compression_Fracture | 0.18 | 0.7 |
| Presence_Acute_Rib_Fracture | 0.16 | 0.6 |
| Presence_Pneumonia | 0.33 | 0.54 |
| Presence_Pneumoperitoneum | 0.42 | 0 |
| Pulmonary_Calcification | 0.49 | 0 |
| Subcutaneous_emphysema | 0.48 | 0 |
| Volume_Pericardial_Effusion_on_CCTA | 0.99 | 0 |

Table A.6: Comparative Accuracies of GPT-4-System and RadLing-System F1 score across each of the 44 pre-defined features from each report

## Section A.7: Working of RadLing-System on a report

We will use the example report mentioned in Table A.9 to show how RadLing-System works.

The first step is Preprocessing and extracting the lines of the report. Here we would only focus on the Findings section.  We run the algorithm sequentially through the sentences to extract the values of the CDEs.

FINDINGS: DEVICES: There are no tubes or lines present. CARDIAC: The cardiac silhouette is normal in size and shape. MEDIASTINUM: Mediastinal and hilar contours are within normal limits. LUNGS/PLEURA: The lungs are clear. There is no pleural effusion. There is no pneumothorax. BONES: There are no acute osseous changes. UPPER ABDOMEN: The portions of the upper abdomen included in this study are within normal limits. OTHER: No other significant abnormalities are identified.

*Sentence 1: DEVICES: There are no tubes or lines present.*

Our retrieval algorithm selects the feature class as Tubes. Type of Tubes selected are: Endotracheal tube, Central venous catheter, Pulmonary artery catheter, Nasogastric tube, Airway stent. The Presence attribute maps to Absent. Placement of Enteric tube is mapped to Unspecified.

*Sentence 2. CARDIAC: The cardiac silhouette is normal in size and shape.*

Feature class selected is Cardiomegaly. Presence attribute gets mapped to Absent.

*Sentence 3. MEDIASTINUM: Mediastinal and hilar contours are within normal limits.*

Feature class selected are Cardiomegaly, Pulmonary calcification, Hilar Enlargement, Scoliosis, Lymphadenopathy, and Pneumothorax. In this case, although cardiomegaly is chosen again, the lexical and semantic similarity of this sentence with cardiomegaly (=0.91) is lower than the previous (=0.98). Hence this value is ignored. For each of the other classes, this is the only occurrence until now, and the Presence attribute is set to Absent.

*Sentence 4. PLEURA: The lungs are clear.*

Feature class selected are Pulmonary Nodule, Airspace consolidation, Atelectasis, Fibrosis, Emphysema and Pleural effusion. The Presence attribute is set to Absent for all of them.

*Sentence 5. There is no pleural effusion.*

Pleural effusion is selected as Feature class, and Presence attribute is set to Absent.

*Sentence 6. BONES: There are no acute osseous changes.*

Feature classes selected are Tubes and Fractures. To note here is that the classes Scoliosis and Fibrosis are also selected as possible classes only with lexical similarity but they are ruled out with semantic similarity. The Presence attribute is set to Absent.

*Sentence 7. UPPER Abdomen: The portions of the upper abdomen included in this study are within normal limits.*

No feature classes are selected.

*Sentence 8. OTHER: No other significant abnormalities are identified.*

No feature classes are selected.

Finally these features selected for the report are mapped to corresponding CDEs. All the other CDEs get corresponding Presence features as unspecified and all other categorical/numerical features are set to the default values of the CDEs.

**Section A.8: Example prompt to GPT-4 System**

*Report:*

DEVICES: There are no tubes or lines present. CARDIAC: The cardiac silhouette is normal in size and shape. MEDIASTINUM: Mediastinal and hilar contours are within normal limits. LUNGS/PLEURA: The lungs are clear. There is no pleural effusion. There is no pneumothorax. BONES: There are no acute osseous changes. UPPER abdOMEN: The portions of the upper abdomen included in this study are within normal limits.OTHER: No other significant abnormalities are identified.

*Prompt to GPT-4 for Feature Extraction using Dynamic Few-Shot:*

'You are radiology assistant. Giving you a chest xray report. Extract following features. Stick to key:value format. For pulmonary nodules, just report size in mm of largest nodule. Say absent only if report says it is absent. Else say unspecified.The keys and their valueset are:Cardiomegaly_Cardiomegaly:[\'absent\', \'present\', \'unspecified\'],Presence_Pleural_Effusion:[\'present\', \'absent\', \'indeterminant\', \'unspecified\'],Lymphadenopathy_COVID-19_on_Chest_CT:[\'present\', \'absent\', \'unspecified\'],Ground_glass_opacity_COVID-19_on_Chest_CT:[\'present\', \'absent\', \'unspecified\'],Air_space_consolidation_COVID-19_on_Chest_CT:[\'absent\', \'present\', \'unspecified\'],Presence_Pneumothorax:[\'absent\', \'present\', \'unspecified\'],Volume_Pericardial_Effusion_on_CCTA:[\'Float with unit-ml\', \'unspecified\'],Pulmonary_edema_Pneumonia:[\'absent\', \'present\', \'unspecified\'],Endotracheal_tube_Chest_Radiograph_Lines_and_Tubes:[\'absent\', \'adequately positioned\', \'malpositioned\', \'unspecified\', \'suboptimally positioned\', \'unspecified\'],Central_venous_catheter_Chest_Radiograph_Lines_and_Tubes:[\'absent\', \'adequately positioned\', \'malpositioned\', \'unspecified\', \'suboptimally positioned\', \'unspecified\'],Pulmonary_artery_catheter_Chest_Radiograph_Lines_and_Tubes:[\'absent\', \'adequately positioned\', \'malpositioned\', \'unspecified\', \'suboptimally positioned\', \'unspecified\'],Nasogastric_tube_Chest_Radiograph_Lines_and_Tubes:[\'absent\', \'adequately positioned\', \'malpositioned\', \'unspecified\', \'suboptimally positioned\', \'unspecified\'],Airway_stent_Chest_Radiograph_Lines_and_Tubes:[\'absent\', \'adequately positioned\', \'malpositioned\', \'unspecified\', \'suboptimally positioned\', \'unspecified\'],Placement_Placement_of_Enteric_Tube:[\'correct\', \'misplaced\', \'needs to be advanced\', \'unspecified\'],Presence_Chest_Radiograph_Pulmonary_Nodules:[\'absent\', \'present\', \'single\', \'multiple\', \'unspecified\'],Presence_Pneumonia:[\'present\', \'absent\', \'unspecified\'],Type_Acute_Clavicle_Fracture:[\'proximal to sternum\', \'proximal to AC joint\', \'middle of the bone, between the sternum and AC joint\', \'unspecified\', \'absent\'],Presence_Acute_Humerus_Fracture:[\'present\', \'absent\',

\'unspecified\'],Presence_Acute_Rib_Fracture:[\'present\', \'absent\',

\'unspecified\'],Fracture_Chronic_Rib_Fracture:[\'absent\', \'present\',

\'unspecified\'],Presence_Vertebral_Compression_Fracture:[\'present\', \'absent\',

\'unspecified\'],Pleural_effusion_size_COVID-19_on_Chest_CT:[\'unspecified\', \'small\', \'medium\',

\'large\', \'unspecified\'],Pleual_effusion_laterality_SHS_Chest_XR:[\'unspecified\', \'absent\', \'right\',

\'left\', \'bilateral\', \'unspecified\'],Pleual_effusion_severity_SHS_Chest_XR:[\'unspecified\', \'absent\',

\'mild\', \'moderate\', \'severe\', \'unspecified\'],Cardiomegaly_severity_SHS_Chest_XR:[\'absent\',

\'unspecified\', \'mild\', \'moderate\', \'severe\'],Size_Pneumothorax:[\'small\', \'medium\', \'large\',

\'absent\', \'unspecified\'],Side_Pneumothorax:[\'right\', \'left\', \'bilateral\', \'absent\',

\'unspecified\'],Lung_tissue_collapse_Pneumothorax:[\'complete\', \'partial\', \'unspecified\',

\'absent\'],Lung_tissue_involvement_Pneumonia:[\'segmental\', \'patchy\', \'lobar\', \'multilobar\',

\'diffuse\', \'cavitary\', \'nodular\', \'unspecified\', \'absent\',

\'unspecified\'],Composition_Pulmonary_Nodule:[\'solid\', \'ground glass\', \'part-solid\', \'fat density\',

\'calcification\', \'cavitation\', \'cystic lucencies\', \'air bronchgrams\', \'unspecified\',

\'absent\'],Size_mm_Pulmonary_Nodule:[\'Float with unit-mm\',

\'unspecified\'],Location_Pulmonary_Nodule:[\'unspecified\', \'left lung\', \'left upper lobe\', \'lingula\',

\'left lower lobe\', \'right lung\', \'right upper lobe\', \'right middle lobe\', \'right lower lobe\',

\'unspecified\', \'bilateral\', \'absent\'],Atelectasis_SHS_Chest_XR:[\'absent\', \'unspecified\', \'present\',

\'unspecified\'],Fibrosis_SHS_Chest_XR:[\'absent\', \'present\',

\'unspecified\'],Pulmonary_Calcification_SHS_Chest_XR:[\'absent\', \'present\',

\'unspecified\'],Hilar_Enlargement_SHS_Chest_XR:[\'absent\', \'present\',

\'unspecified\'],Scoliosis_SHS_Chest_XR:[\'absent\', \'present\',

\'unspecified\'],Eventration_SHS_Chest_XR:[\'absent\', \'present\',

\'unspecified\'],Hernia_SHS_Chest_XR:[\'absent\', \'present\',

\'unspecified\'],Pleural_thickening_SHS_Chest_XR:[\'absent\', \'present\',

\'unspecified\'],Tracheal_Deviation_SHS_Chest_XR:[\'absent\', \'present\',

\'unspecified\'],Subcutaneous_emphysema_SHS_Chest_XR:[\'absent\', \'present\',

\'unspecified\'],Presence_Pneumoperitoneum:[\'present\', \'absent\',

\'unspecified\'],Presence_Thoracic_Pulmonary_Parenchymal_Emphysema:[\'absent\', \'present\',

\'unspecified\'],For sentences in report that are in following list, use the key,value pairs from list. If

sentence not in list, do it yourself :["There are no tubes or lines

present;{\'Endotracheal_tube_Chest_Radiograph_Lines_and_Tubes\': \'absent\',

\'Central_venous_catheter_Chest_Radiograph_Lines_and_Tubes\': \'absent\',

\'Pulmonary_artery_catheter_Chest_Radiograph_Lines_and_Tubes\': \'absent\',

\'Nasogastric_tube_Chest_Radiograph_Lines_and_Tubes\': \'absent\',

\'Airway_stent_Chest_Radiograph_Lines_and_Tubes\': \'absent\',

\'Placement_Placement_of_Enteric_Tube\': \'unspecified\'}", "The cardiac silhouette is normal in size

and shape;{\'Cardiomegaly_Cardiomegaly\': \'absent\', \'Cardiomegaly_severity_SHS_Chest_XR\':

\'absent\'}", "Mediastinal and hilar contours are within normal

limits;{\'Hilar_Enlargement_SHS_Chest_XR\': \'absent\'}", "The lungs are clear. There is no infiltrate or

consolidation.;{\'Air_space_consolidation_COVID-19_on_Chest_CT\': \'absent\'}", "There is no pleural

effusion;{\'Presence_Pleural_Effusion\': \'absent\', \'Pleural_effusion_size_COVID-19_on_Chest_CT\':

\'unspecified\'}", "There is no pneumothorax;{\'Presence_Pneumothorax\': \'absent\',

\'Size_Pneumothorax\': \'absent\'}", "There are no acute osseous

changes.;{\'Endotracheal_tube_Chest_Radiograph_Lines_and_Tubes\': \'absent\',

\'Central_venous_catheter_Chest_Radiograph_Lines_and_Tubes\': \'absent\',

\'Pulmonary_artery_catheter_Chest_Radiograph_Lines_and_Tubes\': \'absent\',

\'Nasogastric_tube_Chest_Radiograph_Lines_and_Tubes\': \'absent\',

\'Airway_stent_Chest_Radiograph_Lines_and_Tubes\': \'absent\',

\'Placement_Placement_of_Enteric_Tube\': \'unspecified\'}

**Section A.9: Technical Glossary**

- Domain-Adapted: Refers to customizing a model to improve its performance on data specific to a particular field or subject area, such as radiology.

- Encoder-Based BERT Language Model: A deep learning model designed to understand the context of words in text by processing data in two directions (encoder). BERT models are trained on large text corpora to generate a contextual understanding of language.

- Parsing: The process of analyzing text to determine its grammatical structure, helping the model to understand the context and meaning of sentences.

- F1 Score: A statistical measure used to evaluate the accuracy of a model by considering both precision (the correctness of positive predictions) and recall (the model's ability to find all positive instances).

- ELECTRA: A model that uses a generator and discriminator approach to more efficiently learn from text data. It is designed to understand and generate text based on the context provided.

- Named Entity Recognition (NER): A process by which an algorithm identifies and categorizes key information in text into predefined categories, such as names of people, organizations, or locations.

- Sentence Transformer (SBERT): An adaptation of the traditional BERT model to produce representations of entire sentences.

- Semantic Similarity and Natural Language Inference: Techniques used to assess the meaning of sentences. Semantic similarity measures how similar two sentences are, while natural language inference determines if a statement logically follows from another.

- Siamese and Triplet Networks: Neural network architectures used to learn from comparisons. They are designed to understand the similarity or difference between pairs of inputs.

- Lexical Similarity and BM25: Methods for assessing how similar two pieces of text are based on their word content. BM25 is a specific algorithm used for ranking documents based on the search query's words.

- Dependency Tree Parsing: A method for analyzing the grammatical structure of a sentence that identifies relationships between words, helping to understand the sentence's meaning more deeply.

- RadLex: A comprehensive lexicon of radiology terms designed to standardize the terminology used in radiology reports and facilitate their analysis and interpretation.

- Dynamic few-shot examples: A technique where a model is given a few examples to learn from on the fly, adapting its responses based on this limited input.

- Embeddings: A way of representing words or phrases as vectors in a high-dimensional space, allowing the model to understand similarity and relatedness between terms.

- Parameter: In the context of machine learning models, a parameter is an internal configuration variable whose value can be estimated from the data.

- Transformer-based architecture: A modern approach in deep learning that models relationships among words (or tokens) in a sentence, regardless of their positions.

- Pretraining and fine-tuning: The process of first training a model on a large corpus of data (pretraining) and then adjusting (fine-tuning) it on a smaller, domain-specific dataset to specialize its capabilities.

- Cross encoder vector comparison: A method of comparing the embeddings (vector representations) of sentences or documents to determine their similarity or relevance.

- Cosine similarity: A metric used to measure how similar two vectors are in their orientation in a multi-dimensional space, often used to determine the similarity between two pieces of text.